# Accenture-NVS1: A Novel View Synthesis Dataset

Thomas Sugg[1], Kyle O'Brien[1], Lekh Poudel[1], Alex Dumouchelle[1], Michelle Jou[1], Marc Bosch[1*]
Deva Ramanan[2], Srinivasa Narasimhan[2], Shubham Tulsiani[2]

[1]Accenture Federal Services [2]Carnegie Mellon University

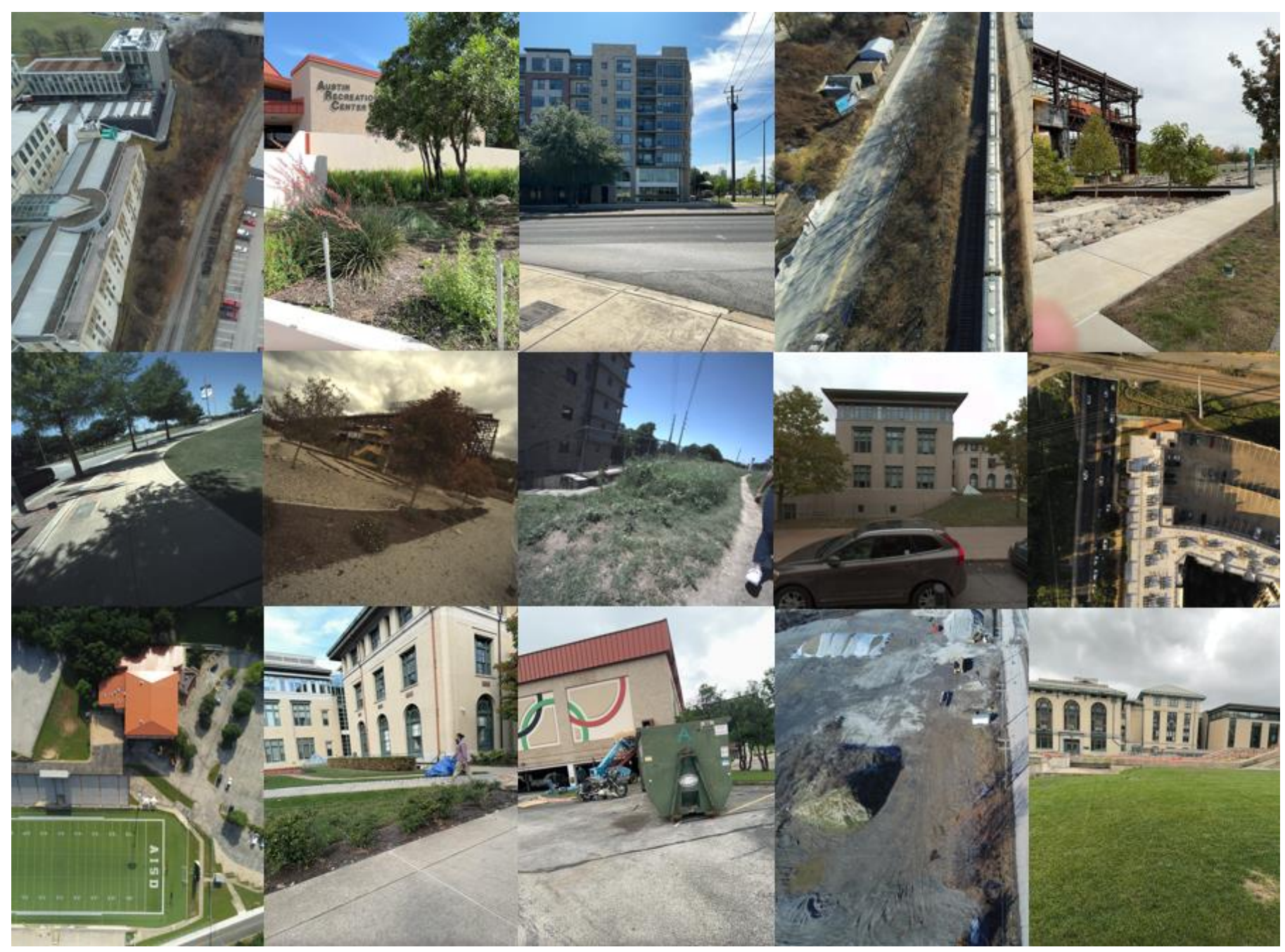

*Figure 1: The ACC-NVS1 Dataset is a real-world scene dataset of six unique environments with 148,000 images. These images are captured from varying altitudes and camera sensors.*

## ABSTRACT

This paper introduces ACC-NVS1, a specialized dataset designed for research on Novel View Synthesis specifically for airborne and ground imagery. Data for ACC-NVS1 was collected in Austin, TX and Pittsburgh, PA in 2023 and 2024. The collection encompasses six diverse real-world scenes captured from both airborne and ground cameras, resulting in a total of 148,000 images. ACC-NVS1 addresses challenges such as varying altitudes and transient objects. This dataset is intended to supplement existing datasets, providing additional resources for comprehensive research, rather than serving as a benchmark.

## 1. INTRODUCTION

Novel view synthesis (NVS) has emerged as a pivotal application in computer vision and graphics, particularly with the advent of Radiance Field methods like 3D Gaussian Splatting [1] and Neural Radiance Fields [2]. These advancements drive the need for comprehensive datasets. While numerous datasets exist for objects and object-centric scenes, like CO3D [3], Mip-NeRF 360 [4], and Tanks and Temples [5], there is a growing demand for datasets that encompass larger environments, such as real-world buildings, which pose unique challenges and opportunities for NVS research.

---

* Corresponding author: marc.bosch.ruiz@afs.com

Crowd-sourced and web-scraped images offer substantial benefits for creating diverse and large-scale datasets. However, obtaining airborne images of these scenes and accurately calibrating them, especially when they are not well-known landmarks, remains a challenge due to coordination between airborne and ground data collections. This scarcity of airborne imagery limits the ability of existing datasets to fully represent complex real-world scenarios, leading to a domain gap when models trained solely on ground perspectives are applied to use cases that include both airborne and ground perspectives. Due to the comprehensive nature of collecting airborne and ground imagery simultaneously, or because of the downstream tasks that models trained on those datasets perform, most scene-level multi-view datasets only collect one perspective, for example, UrbanScene3D [7], Quad-6K [12], and Mill 19 [10] only contain airborne imagery and KITTI [8] and Block-NeRF [13] contain ground imagery.

To address these limitations, we introduce ACC-NVS1, a dataset specifically designed to enhance research in NVS and other vision tasks like feature matching, 3D geometry and scene reconstruction. Data for ACC-NVS1 was collected in Austin Texas and Pittsburgh Pennsylvania, capturing six diverse real-world scenes through both airborne and ground cameras, resulting in a total of 148,000 images. Our dataset addresses challenges such as varying altitudes and transient objects, providing a valuable resource for testing and improving the performance of state-of-the-art models on real-world data. The ACC-NVS1 dataset is not intended to serve as a benchmark but rather to supplement existing datasets, offering complementary data for comprehensive research. ACC-NVS1 aims to advance the field of 3D vision, enabling more robust and generalizable models.

## 2. RELATED WORK

**Multi-view Datasets.** It is rare to find large-scale datasets for NVS where all scenes have paired ground and airborne imagery. Recently, DL3DV-10K [6] introduced over 10,000 diverse scenes for NVS with 1% of those being drone captures. Although DL3DV-10K serves as a rich resource for training NVS models, it does not contain scenes with paired airborne and ground cameras. MegaScenes [9] is a multi-view dataset with 430K scenes sourced from Wikimedia Commons and contains imagery from both airborne and ground cameras. However, the camera poses and camera intrinsics are estimated using COLMAP [11]. Since the scenes are calibrated using COLMAP, there may be inconsistencies in scene coverage across all examples and camera poses may be inaccurate.

To address the limited availability of airborne-ground pairs in large-scale datasets, we offer ACC-NVS1, a focused multi-view dataset with dense coverage from both ground and airborne imagery, as an additional resource for NVS model training. Additionally, one of the scenes provided in ACC-NVS1 is of Mill 19 in Pittsburgh that was previously collected for Mega-NeRF [10]. Our dataset improves upon their 1,809 drone images with even more drone imagery and tens of thousands of ground images.

## 3. ACC-NVS1 DATASET

Our objective for ACC-NVS1 was to collect extensive, high-quality scenes that accurately represent real-world environments from airborne and ground perspectives. Data collections were conducted similarly for all sites by collecting multiple trajectories around the site with and without transients. Data was collected in various navigable environments that include structures and fixed objects such as buildings, sidewalks, roadways, stairs, ramps, light poles, signs, vegetation, etc. Data collection was conducted with RGB Imagery and LiDAR Multispectral using a combination of sensors and platforms including unmanned aerial vehicles (drones), ground vehicle based mobile mapping, and other handheld imaging devices. For the collection in Pittsburgh, PA ("CMU"), we captured three sites on the Carnegie Mellon University campus during the winter season. For the collection in Austin, TX ("ATX"), we captured two different sites in the downtown area during the spring season. Data collection for each city took approximately a week. A summary of image counts is visible on Table 1 for each scene.

| Site Name | Number of Images |
|---|---|
| **CMU-mall** | 17,686 |
| **CMU-fine-arts** | 18,756 |
| **CMU-mill-19** | 32,499 |
| **ATX-gables-park** | 53,941 |
| **ATX-rec** | 25,509 |
| **Total** | **148,391** |

*Table 1: Number of images captured per site.*

All images are provided with metadata that includes camera positions and camera intrinsics. The images were calibrated and geolocated through mapping software, and the camera geolocation relied on multiple techniques like using ground control points, RTK, PPK, Lidar, and GPS.

**Sensors.** Four different types of sensors were used to capture the imagery. iPhone and Drone camera sensors were selected to capture RGB images while the Backpack- and Truck-mounted sensors were used to capture RGB and LiDAR data. The following table summarizes details about each acquisition sensor and displays sample imagery from each.

| Sensors | Description | Resolution | Image Examples |
|---|---|---|---|
| **iPhone** | A handheld camera in an iPhone, attached to the Pix4D viDoc hardware (for camera location and pose information) | 1440x1920 | |
| **Drone** | 21 MP wide-angle camera with 32x zoom capability | 3456x4608 | |
| **Backpack** | A set of five 4-megapixel cameras mounted on the side of the Pegasus wearable reality capture device | 2048x2048 | |
| **Truck** | A subset of cameras from the 6-lens spherical camera of the truck-mounted Trimble MX50 imagery/lidar scanner | 2048x2048 | |

*Table 2: Acquisition sensor description and examples.*

**Areas of Interest (AOIs).** Each AOI was chosen because of its location inside an urban environment with diverse and unique building architectures, ranging from an old steel mill, a university campus, an apartment complex, and a recreation center. Due to the nature of some of these navigable environments, the truck-mounted system was not used for the Austin AOIs. Figure 2 (right) displays which sensors were used at each AOI.

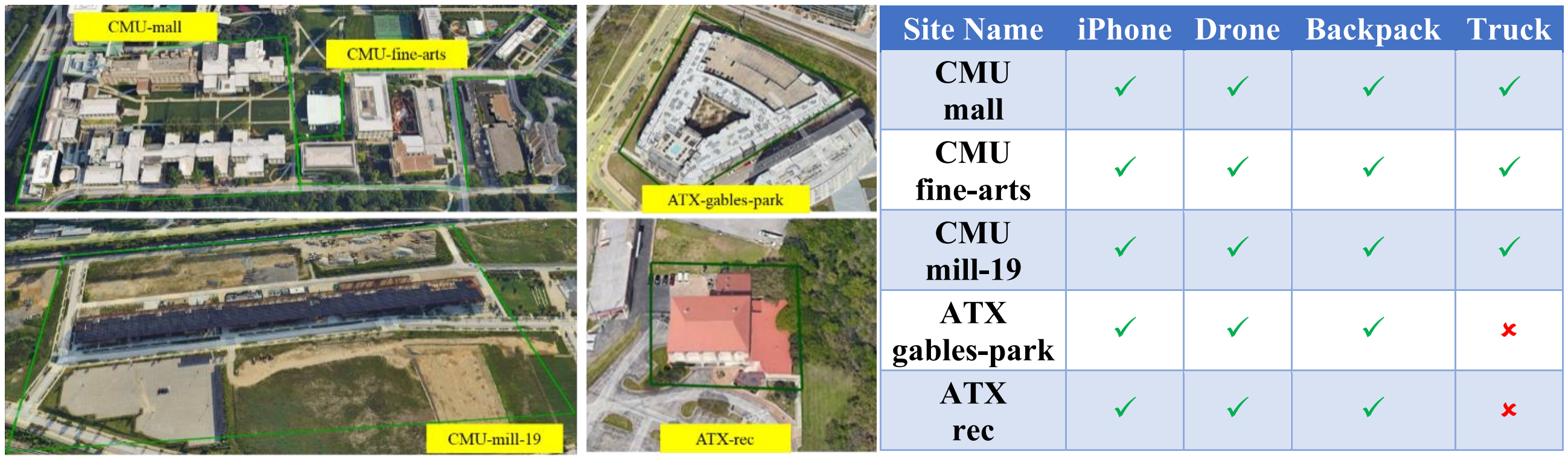

| Site Name | iPhone | Drone | Backpack | Truck |
|---|---|---|---|---|
| CMU mall | ✓ | ✓ | ✓ | ✓ |
| CMU fine-arts | ✓ | ✓ | ✓ | ✓ |
| CMU mill-19 | ✓ | ✓ | ✓ | ✓ |
| ATX gables-park | ✓ | ✓ | ✓ | × |
| ATX rec | ✓ | ✓ | ✓ | × |

*Figure 2: Satellite images of all ACC-NVS1 scenes with bounding box polygons showing the scene's extent (left). Sensors used at each AOI. (right)*

**Transients**. A transient occlusion refers to a temporary obstruction or blockage of a particular object or area within an image. It occurs when an object or part of an object is momentarily hidden or obscured from view due to factors such as movement, interference, or other objects passing in front of it. For example, if a person is walking by or a vehicle drives past the camera, objects or scenes may be transiently occluded as the camera's viewpoint changes or due to objects entering the frame. While the presence of transients often impacts the quality of the novel view renders, it is a realistic capture condition for these types of novel view synthesis algorithms to overcome. As a result, to enable better real-world simulation, ACC-NVS1 included purposeful occlusions of objects using tarps, people passing in front of the sensors, and fingers over the lens (iPhone camera only), as shown in Figure 3. Each trajectory was performed with and without transient occlusions.

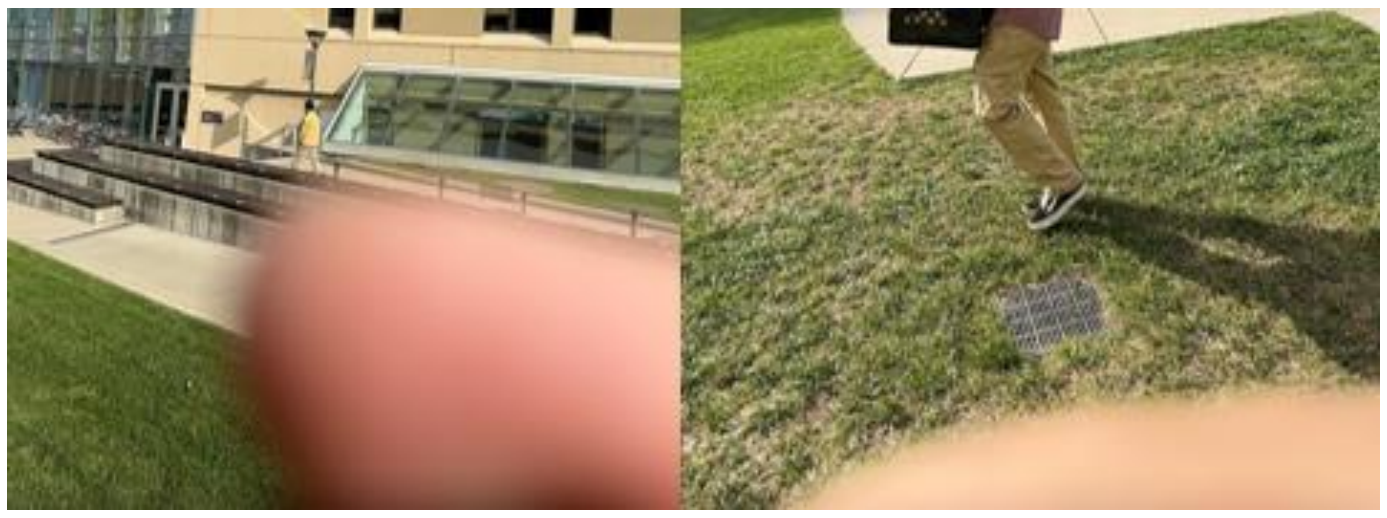

*Figure 3: Examples of transient artifacts.*

**Post processing.** To comply with technical and legal safeguards, all collected images have been post processed to scrub Personally Identifiable Information (PII) from the raw images. PII refers to information that can be used to identify an individual uniquely which incorporates two main categories: faces and license plates. To remove PII from the data, an automated pipeline was used to (1) detect faces and license plates then (2) blur all identified regions in the images as shown in Figure 4. However, since the detection algorithms aren't always accurate, a manual review process was put in place to confirm the presence of PII. PII scrubbing was conducted on local servers and reviewed by trained staff.

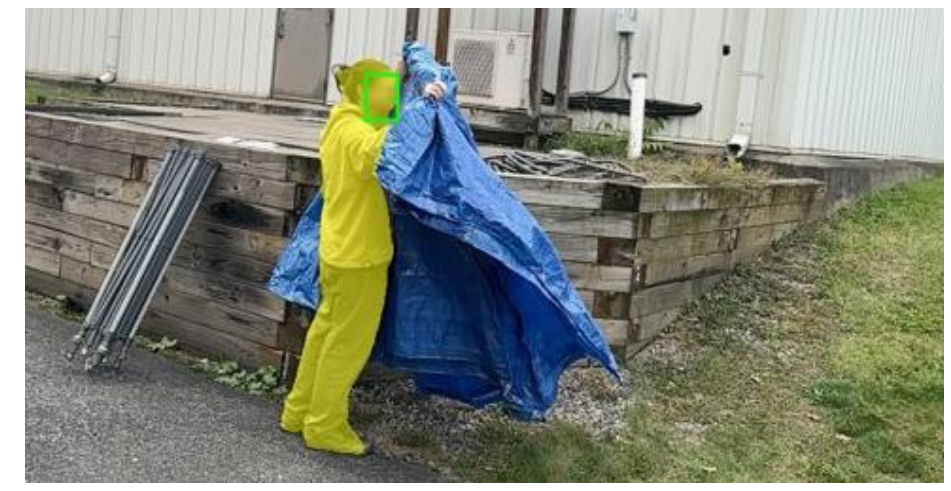

*Figure 4: Example of PII detection and removal.*

## 4. APPLICATIONS

ACC-NVS1 can be useful for multiple downstream applications in novel view synthesis. This section details the different use cases the dataset benefits research and experimentation.

**Finger Blur Detection.** ACC-NVS1 contains thousands of images with transient objects like people, vehicles, and fingers over the lens. Detecting and segmenting such artifacts can assist other downstream operations in NVS training and/or rendering such as omitting the artifact image or mitigating the occluded regions via inpainting. For example, Figure 5 shows the results of a custom finger blur detector we developed using ACC-NVS1. Detection of this class of artifact is uncommon in research literature.

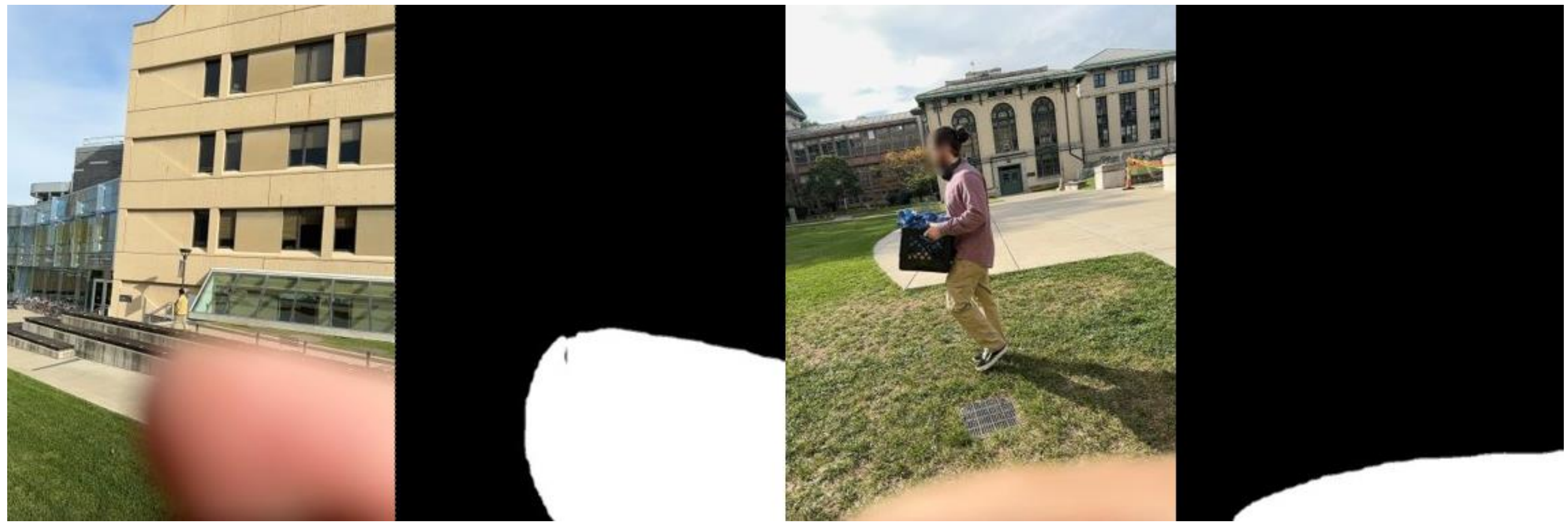

*Figure 5: Predictions from a custom finger blur detection model.*

**Reconstruction.** The starting point for many NVS applications is a collection of uncalibrated images. Structure-from-Motion (SfM) is commonly used to estimate camera poses and intrinsics. In Figure 6 we show COLMAP outputs using a small sample of images from two ACC-NVS1 scenes.

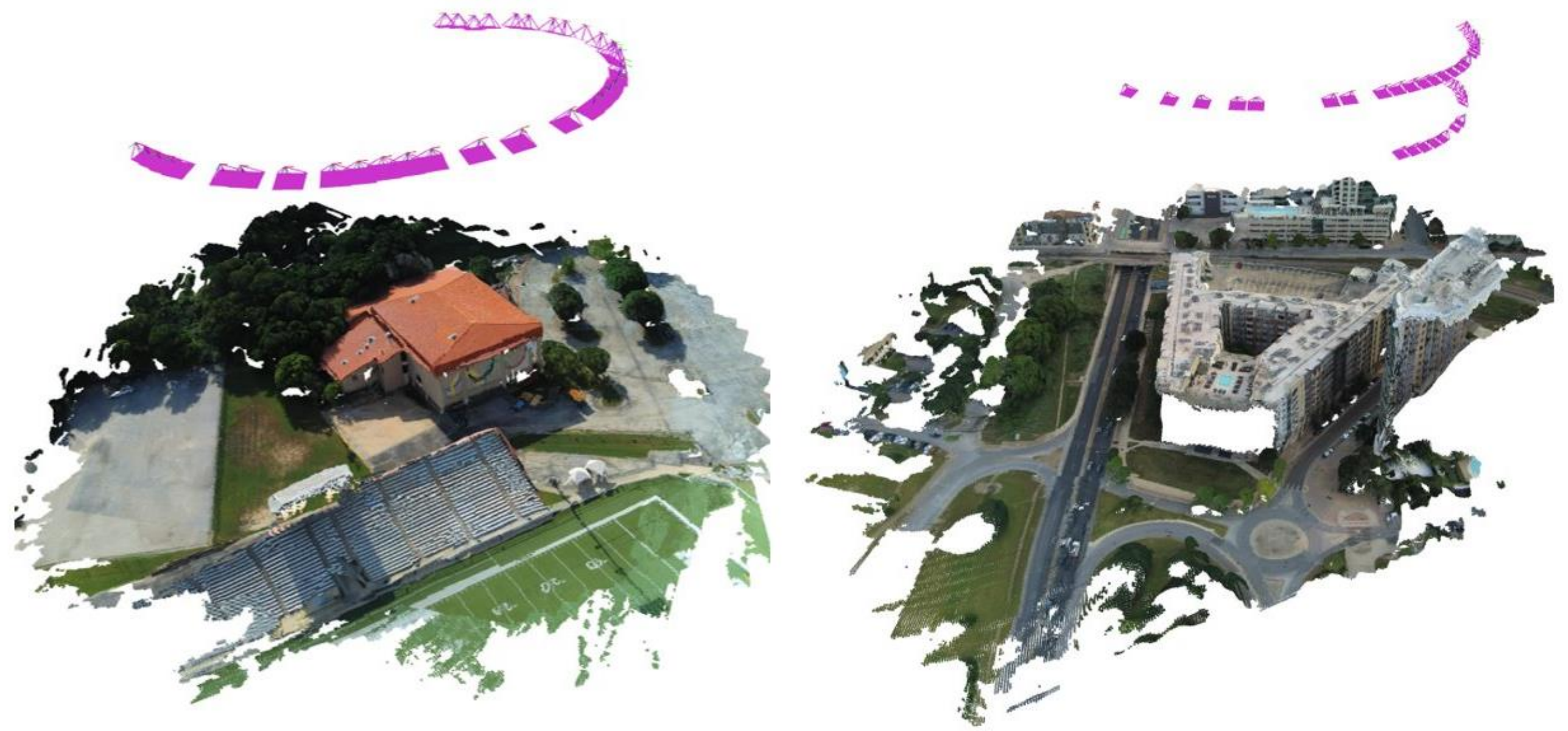

*Figure 6: Outputs from Structure-from-Motion.*

**Monocular Depth Estimation and Projection.** Some surfaces and parts of scenes are still incomplete after SfM, so Multi-View Stereo (MVS) is an option to make a denser reconstruction. However, this is time-consuming and may still result in missing surfaces and empty parts of a scene. For NVS techniques like 3D Gaussian Splatting that rely on a point cloud for initialization, we have experimented with using monocular depth estimators like DepthAnythingV2 [14] to fill in those missing areas accurately and quickly. We do this by aligning the relative depth from DepthAnythingV2 with the sparse points from SfM. Figure 7 shows a couple depth predictions and the final point cloud after projecting the monocular depth predictions in the COLMAP scale.

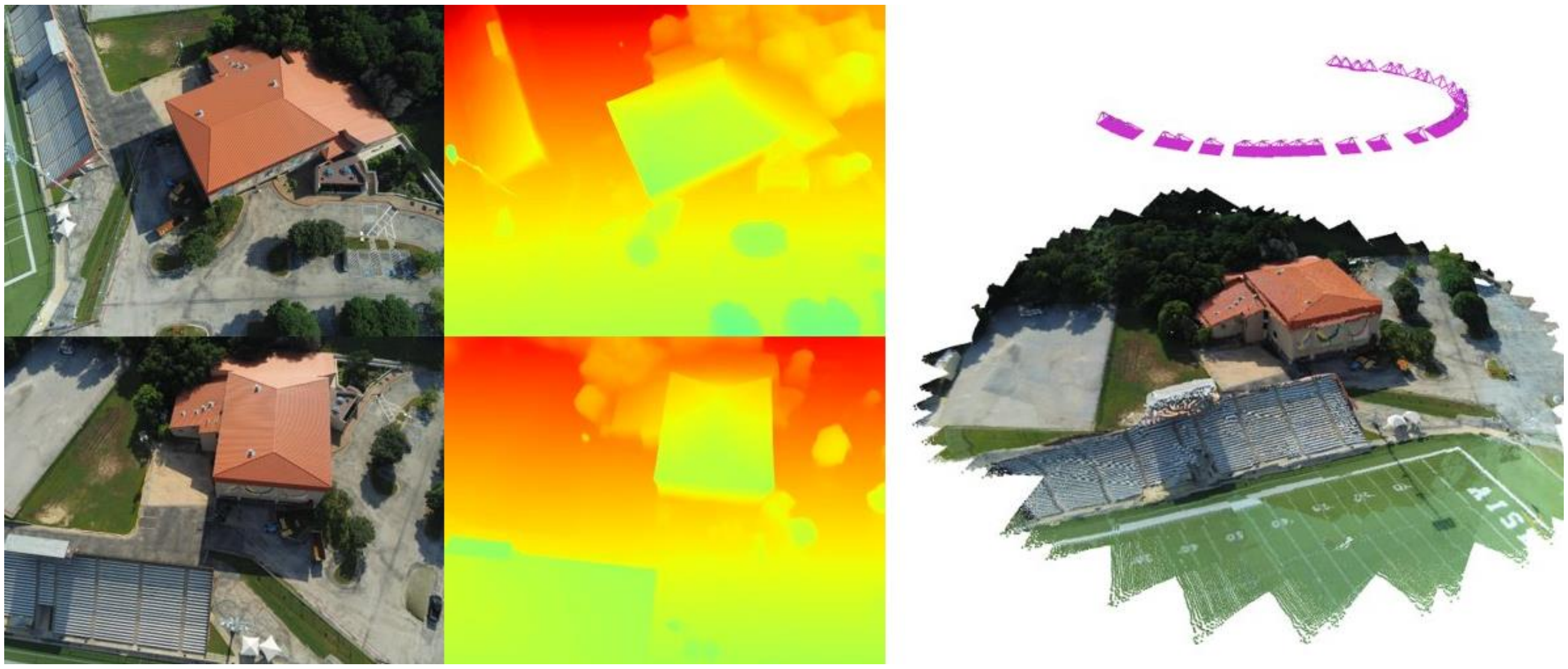

*Figure 7: DepthAnythingV2 predictions projected in COLMAP scale.*

## 5. CONCLUSION

We introduce ACC-NVS1, a multi-view scene dataset created by capturing images from varying altitude across six real-world environments. We propose ACC-NVS1 as an additional resource for training models for 3D vision tasks such as Novel View Synthesis, feature matching, scene reconstruction and/or camera calibration.

## ACKNOWLEDGEMENTS

We would like to thank the CMU team (Khiem Vuong, Jeff Tan, Arun Vasudevan, and Robert Tamburo) for helping us coordinate collections for the CMU sites and the Voxelmaps team who collected the drone and LiDAR imagery with us onsite. We would also like to thank Accenture teammates (Peter Rasmussen, Danielle Catalfamo, Lindsey Castin, Adam Stevens, Michelle Rybak, Howard Brand, Jiong Huang, and Nishant Gurrapadi) who supported the PII scrubbing effort and onsite collections with Voxelmaps.

This work was supported by the Intelligence Advanced Research Projects Activity (IARPA) via Department of Interior/ Interior Business Center (DOI/IBC) contract number 140D0423C0074. The U.S. Government is authorized to reproduce and distribute reprints for Governmental purposes notwithstanding any copyright annotation thereon. Disclaimer: The views and conclusions contained herein are those of the authors and should not be interpreted as necessarily representing the official policies or endorsements, either expressed or implied, of IARPA, DOI/IBC, or the U.S. Government.

## REFERENCES

[1] Kerbl, Bernhard, et al. "3d gaussian splatting for real-time radiance field rendering." *ACM Trans. Graph.* 42.4 (2023): 139-1.

[2] Mildenhall, Ben, et al. "Nerf: Representing scenes as neural radiance fields for view synthesis." *Communications of the ACM* 65.1 (2021): 99-106.

[3] Reizenstein, Jeremy, et al. "Common objects in 3d: Large-scale learning and evaluation of real-life 3d category reconstruction." *Proceedings of the IEEE/CVF international conference on computer vision*. 2021.

[4] Barron, Jonathan T., et al. "Mip-nerf 360: Unbounded anti-aliased neural radiance fields." *Proceedings of the IEEE/CVF conference on computer vision and pattern recognition*. 2022.

[5] Knapitsch, Arno, et al. "Tanks and temples: Benchmarking large-scale scene reconstruction." *ACM Transactions on Graphics (ToG)* 36.4 (2017): 1-13.

[6] Ling, Lu, et al. "Dl3dv-10k: A large-scale scene dataset for deep learning-based 3d vision." *Proceedings of the IEEE/CVF Conference on Computer Vision and Pattern Recognition*. 2024.

[7] Lin, Liqiang, et al. "Capturing, reconstructing, and simulating: the urbanscene3d dataset." *European Conference on Computer Vision*. Cham: Springer Nature Switzerland, 2022.

[8] Andreas Geiger, et al. "Vision meets Robotics: The KITTI Dataset". *International Journal of Robotics Research (IJRR)*. (2013).

[9] Tung, Joseph, et al. "Megascenes: Scene-level view synthesis at scale." *European Conference on Computer Vision*. Cham: Springer Nature Switzerland, 2024.

[10] Turki, Haithem, Deva Ramanan, and Mahadev Satyanarayanan. "Mega-nerf: Scalable construction of large-scale nerfs for virtual fly-throughs." *Proceedings of the IEEE/CVF conference on computer vision and pattern recognition*. 2022.

[11] Schonberger, Johannes L., and Jan-Michael Frahm. "Structure-from-motion revisited." *Proceedings of the IEEE conference on computer vision and pattern recognition*. 2016.

[12] Crandall, David, et al. "Discrete-continuous optimization for large-scale structure from motion." *CVPR 2011*. IEEE, 2011.

[13] Tancik, Matthew, et al. "Block-nerf: Scalable large scene neural view synthesis." *Proceedings of the IEEE/CVF conference on computer vision and pattern recognition*. 2022.

[14] Yang, Lihe, et al. "Depth anything v2." *Advances in Neural Information Processing Systems* 37 (2024): 21875-21911.